\documentclass[11pt]{article}

\usepackage[final]{acl}

\usepackage{times}
\usepackage{latexsym}
\usepackage{amsmath}
\usepackage{amssymb}
\usepackage{mathtools}
\usepackage{amsthm}
\usepackage{booktabs}
\usepackage{multirow}
\usepackage{makecell}

\usepackage[T1]{fontenc}

\usepackage[utf8]{inputenc}

\usepackage{microtype}

\usepackage{inconsolata}

\usepackage{graphicx}

\title{Can Large Language Models Reason and Optimize Under Constraints?}

\author{Fabien Bernier$^{1}$, Salah Ghamizi$^{1,2}$, Pantelis Dogoulis$^{1}$, Maxime Cordy$^{1}$ \\
        $^{1}$SnT -- University of Luxembourg, Luxembourg \\
        $^{2}$Luxembourg Institute of Health (LIH), Luxembourg
        }

\begin{document}
\maketitle
\begin{abstract}
Large Language Models (LLMs) have demonstrated great capabilities across diverse natural language tasks; yet their ability to solve abstraction and optimization problems with constraints remains scarcely explored. In this paper, we investigate whether LLMs can reason and optimize under the physical and operational constraints of Optimal Power Flow (OPF) problem. We introduce a challenging evaluation setup that requires a set of fundamental skills such as reasoning, structured input handling, arithmetic, and constrained optimization. Our evaluation reveals that SoTA LLMs fail in most of the tasks, and that reasoning LLMs still fail in the most complex settings. Our findings highlight critical gaps in LLMs' ability to handle structured reasoning under constraints, and this work provides a rigorous testing environment for developing more capable LLM assistants that can tackle real-world power grid optimization problems.
\end{abstract}

\section{Introduction}
Large Language Models (LLMs) have achieved notable performance across a wide range of natural language understanding and generation tasks, from open-ended dialogue and code synthesis to mathematical reasoning and scientific question answering~\cite{guo2025deepseek,srivastava2023beyond,wang2024mmlu}. Yet a critical question remains largely unanswered: \textit{can LLMs reason and optimize under constraints?} Real-world decision-making problems, spanning power grid management, financial operations, and cyber-security, require not only language competence but also the ability to jointly interpret structured inputs, perform multi-step arithmetic, satisfy interacting physical or logical constraints, and converge to feasible, near-optimal solutions. These challenges go far beyond what current benchmarks assess.

The rewards of closing this gap are high. In safety-critical infrastructure such as power grids, for instance, LLMs capable of genuine constrained reasoning could act as world models, avoiding heavy simulations and improving autonomous planning / operational assistance that current systems cannot support. More broadly, advancing LLM competence on structured optimization problems would unlock a new class of AI assistants for high-value engineering and operational domains.

\begin{figure*}[t]
    \centering
    \includegraphics[width=\linewidth]{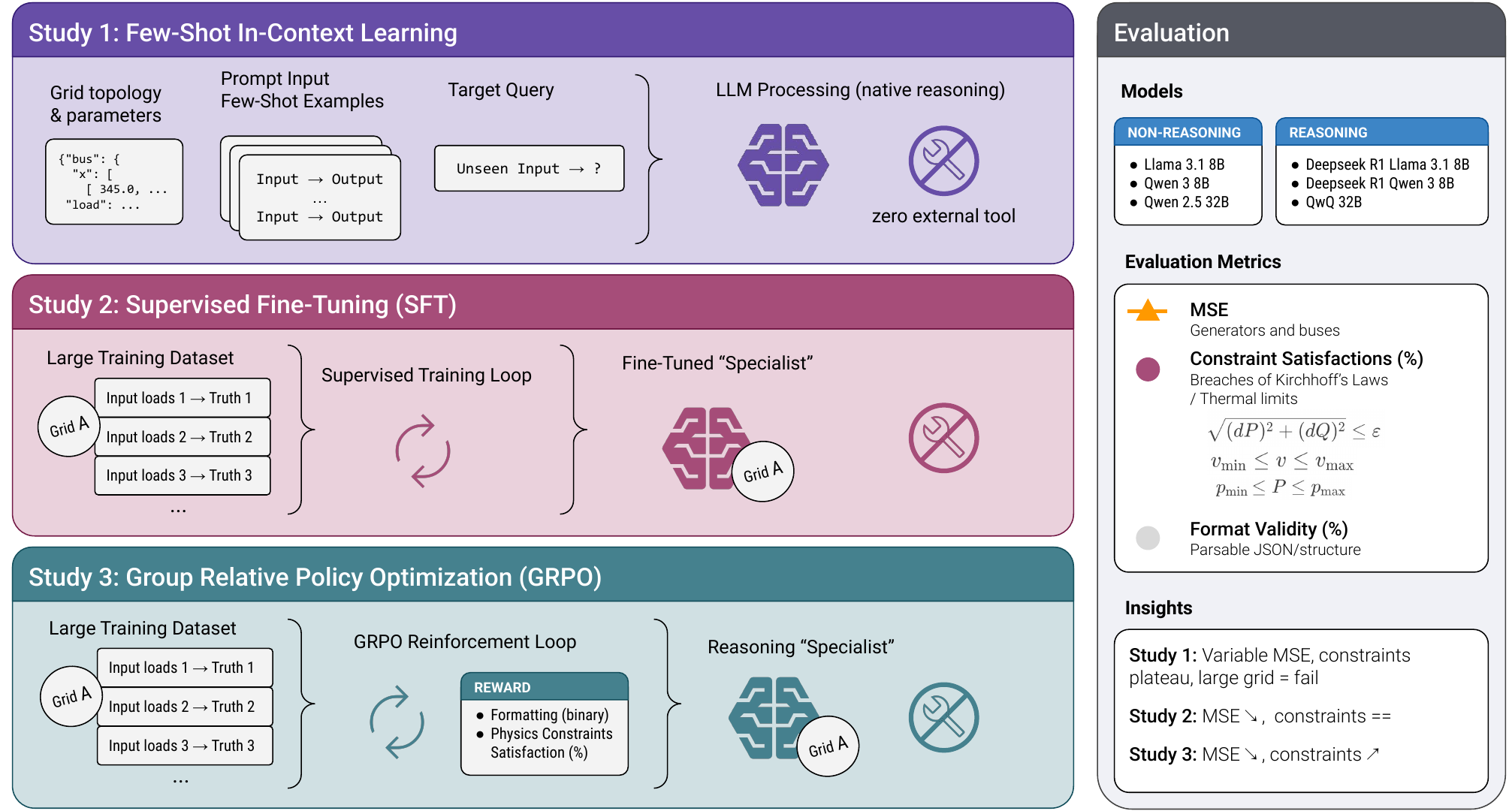}
    \caption{Overview of our study. LLMs are evaluated with simple in-context learning, and in the evaluation settings after supervised fine-tuning (for non-reasoning models) and group relative policy optimization (for reasoning models); all models outputs are evaluated with the same metrics: MSE, constraint satisfaction, and structure validity.}
    \label{fig:studies}
\end{figure*}

Existing benchmarks, however, fall short in evaluating these capabilities in a rigorous and realistic manner. General reasoning benchmarks such as MMLU~\cite{wang2024mmlu} and GPQA~\cite{rein2024gpqa} assess broad knowledge and expert-level question answering, but do not require iterative numerical optimization or constraint satisfaction over structured physical systems. Logical reasoning benchmarks such as ARC-AGI~\cite{chollet2019measureintelligence}, SATBench~\cite{wei2025satbench} and ZebraLogic~\cite{lin2025zebralogic} probe constraint satisfaction in formal or combinatorial settings, but rely on synthetic puzzles disconnected from real-world engineering complexity. Domain-specific work on LLMs for power systems~\cite{huang2024large,ren2025can} or finance has largely focused on agents augmented with external solvers or reinforcement learning, missing the core question of whether the model itself can reason through constraints end to end.

To address this gap, we introduce a new challenging task that evaluates LLMs in performing end-to-end constrained reasoning and numerical optimization over realistic power-grid operation scenarios using well known datasets. To further investigate LLMs abilities on such complex tasks, we assess them extensively after applying supervised fine-tuning (SFT) and group relative policy optimization (GRPO).

Our empirical evaluation of state-of-the-art LLMs, including both instruction-tuned models and reasoning-augmented variants, across small and medium sizes, reveals a consistent failure pattern. Across virtually all tasks requiring genuine optimization under constraints, models remain at a constraint satisfaction rate of approximately 55--60\%, regardless of architecture, scale, or training regime. Reasoning models, despite their extended chain-of-thought generation, do not systematically outperform their non-reasoning counterparts. Supervised fine-tuning improves response formatting but fails to improve physical feasibility, confirming shortcuts in reasoning. Reinforcement learning with constraint-satisfaction rewards, however, yields modest but meaningful improvements on some grid topologies.

In summary, our contributions are as follows: (1) a new task for evaluating LLM reasoning under physical constraints via the OPF problem, (2) a rigorous evaluation protocol with constraint-aware metrics, and (3) an extensive empirical study spanning vanilla, SFT, and GRPO-trained models, revealing a consistent failure plateau in constraint satisfaction.

\section{Related work}

\subsection{LLMs for Power Systems}
The application of LLMs to energy and power systems has seen significant expansion across various sub-domains. 
Within the specific area of optimization, existing literature mainly explores the use of LLMs as high-level agents. For instance, LLM-SUC~\cite{ren2025can} focuses on training LLM-based agents via reinforcement learning to optimize financial costs in energy markets. Specialized frameworks such as LLM4OPF and LLM4EV~\cite{huang2024large} target specific technical challenges in Optimal Power Flow and electric vehicle scheduling. However, these agent-based approaches often rely on reinforcement learning for cost optimization, or external tool calling, rather than direct multi-constraint reasoning.

In simulation tasks, DALINE-GPT~\cite{mirshekali2025review} has demonstrated high success rates in generating simulation code (more specifically for MATPOWER), although it still struggles to construct small distribution networks without intensive prompt engineering. Finally, to move beyond traditional \textit{regression} tasks, the existing work mostly focuses on load forecasting: models like EF-LLM~\cite{qiu2024ef} and LFLLM~\cite{liu2024lfllm} use fine-tuning to handle short-term forecasting across multiple voltage levels. Newer \textit{cross-modal} approaches, such as the GPT-Agent~\cite{yan2024probabilistic} for PV power forecasting, have also integrated linguistic weather data with numerical vectors to improve interpretability and accuracy, but still lack the structured reasoning required to solve multi-constraint optimization problems with interacting variables. Indeed, even recent work that directly targets optimization problems such as OPF~\cite{bernier2025powergraph} evaluates LLM performance solely through MSE, without verifying whether the produced solutions satisfy the underlying physical and operational constraints, letting feasibility guarantees largely unaddressed.

\subsection{LLM Reasoning Capabilities}
Traditional reasoning benchmarks have smoothly transitioned from testing basic linguistic understanding to assessing massive multitask knowledge (e.g. MMLU~\cite{wang2024mmlu}, BIG-bench~\cite{srivastava2023beyond}) and expert-level reasoning (GPQA~\cite{rein2024gpqa}). Despite this evolution, several limitations persist. Logical reasoning benchmarks such as SATBench~\cite{wei2025satbench} and ZebraLogic~\cite{lin2025zebralogic} evaluate constraint-satisfaction abilities, but they often rely on synthetic puzzles or formal logic statements that do not reflect the noise and complexity of real-world engineering tasks. Other benchmarks focus on narrow aspects of reasoning: Planbench~\cite{valmeekam2023planbench} assesses planning and reasoning about change, while Causalbench~\cite{wang2024causalbench} evaluates causal inference but does not address optimization or structured constraint satisfaction.

Recent surveys indicate that while LLMs achieve high scores on static benchmarks, they often rely on shortcut \textit{learning-memorizing} solution templates rather than engaging in generalizable reasoning. This leads to a lack of process credibility, where models may reach the correct answer through flawed logical steps. \cite{guo2025deepseek} highlights a significant divergence between Supervised Fine-Tuning (SFT) and Reinforcement Learning (RL) approaches: while SFT helps models follow specific output formats, it can lead to \textit{pattern matching} that occasionally suppresses pre-trained reasoning logic or fails on out-of-distribution tasks~\cite{chu2025sft}.
State-of-the-art models are additionally fragile when exposed to small perturbations or adversarial inputs, often failing tasks they previously solved if the context is slightly modified~\cite{ni2025survey}. Furthermore, widespread data contamination in training sets often inflates performance metrics, masking the models' true inability to handle novel, out-of-distribution optimization problems~\cite{ni2025survey}.

\subsection{LLMs on Constraint Satisfaction Tasks}

Prior work on LLMs and constrained reasoning can be organized into three main categories. In the first one, works are focused on the problem formulation from natural language. NL4Opt evaluates whether models can recover decision variables, objectives, and constraints from textual optimization descriptions \citep{ramamonjison2023nl4opt}. More recent work in the constraint-programming community extends this direction by asking LLMs to generate executable constraint models from combinatorial problem statements and by introducing broader benchmarks for this capability \citep{michailidis2024constraint,michailidis2025cp}. While this line of work is closely related to optimization, it primarily evaluates modeling and formalization accuracy, rather than whether an LLM can itself carry out constrained search and produce a valid solution end to end.

A second direction of research focuses on direct constraint satisfaction and combinatorial reasoning. The authors in \cite{madusanka2024natural}, examine satisfiability judgments under different logical fragments expressed in natural language, while in LR$^2$Bench \citep{chen2025lr2bench}, the authors evaluate CSP-style tasks that require long reasoning chains, reflection, and backtracking. In a related optimization setting, EHOP \citep{duchnowski-etal-2025-knapsack} shows that LLM performance on NP-hard problems is highly sensitive to surface presentation, suggesting dependence on familiar formulations rather than robust abstraction over constraints. Taken together, these studies suggest that current models remain fragile when constraints interact strongly, search depth increases, or problem formulations deviate from familiar templates.

A third direction is based on the work to improve reliability through neuro-symbolic methods. Logic-LM and SatLM translate natural-language problems into symbolic or declarative representations and then rely on external solvers for inference \citep{pan2023logic,ye2023satlm}. In the constraint-programming literature, LLMs have also been used to generate streamlining constraints that accelerate downstream search \citep{voboril2025generating}. These methods often improve accuracy, but they also move away from fully end-to-end reasoning. In contrast, our work focuses on a demanding task requiring end-to-end model competence: jointly interpreting context, satisfying the constraints, and producing accurate responses in a hard numerical problem.

\section{Method}
We design a rigorous evaluation framework to assess whether LLMs can reason and optimize under the physical and operational constraints of a complex real-world problem.
We study the Optimal Power Flow (OPF) problem, which is a fundamental to operational planning and grid management. This task requires input interpretation, multi-step arithmetic, and simultaneous satisfaction of interacting physical constraints.

\subsection{Problem Formulation of Optimization with Reasoning}

Non-Convex Non-Linear Programming (NLP) problems provide a powerful framework for modeling and evaluating logical reasoning capabilities of the LLMs. In such problems, solutions must satisfy a set of constraints over variables and their possible value, while iteratively converging towards an optimal solution that minimizes an expected reward or fitness function. The OPF problem seeks to minimize a cost function (like generation fuel cost or system losses) subject to the physical laws of the electrical grid. In practice, solving such a problem by an LLM would demonstrate true reasoning capabilities given it requires solving multiple complex, iterative sub-problems:

\paragraph{Abstraction.} Natural language parsing and understanding of a textual power grid description, and a topological representation of its components. The main challenge lies in the capability of the LLM to maintain a "chain of thought" when building its representation. If the logical chain breaks at step two, the final output will fail to represent an OPF solution.

\paragraph{Mathematics.} Translating that description into algebraic equations and inequalities. The main challenges for an LLM lie in identifying the right formulation (complexe AC OPF, or simplified DC OPF approximation) and its ability to reason about the trade-offs between computational efficiency and physical accuracy, and structuring those equations into a syntax that an optimization algorithm can solve.

\paragraph{Multi-Step optimization.} Computing the numerical solution iteratively using well known algorithms such as Newton-Raphson and Interior Point methods. The main challenges are the mathematical bottleneck and the exponential error compounding in iterative methods. For instance, to solve a 9-bus system, the very first step is to construct the Admittance Matrix. For an AC OPF, this is a 9×9 matrix of complex numbers. To solve the power flow equations, algorithms must iteratively invert or factorize these matrices (often building a Jacobian matrix of partial derivatives). LLMs may not reliably perform matrix inversion or solve systems of linear equations without tools, and there is high risk for the LLM to introduce a small rounding error. In the next iteration, that error is fed back into the non-linear equations, compounding exponentially.

Given the complexity of the problem at hand, our study explores to which extent LLMs could approximate the solution, especially when provided examples, and fine-tuned for this task.

\subsection{Power Grid Optimization task}

In this task, LLMs must optimize power generation and distribution across a network while satisfying multiple physical and operational constraints. The input consists of a power grid topology represented as a heterogeneous graph, where nodes represent different components (buses, generators, loads, and transmission lines), each with distinct feature sets. The grid configuration includes active and reactive power demands at load nodes and generator capacity limits.

\paragraph{Dataset.}
We build our dataset using \texttt{PyTorch Geometric} framework and its interface to 
\texttt{OPFData}~\cite{lovett2024opfdatalargescaledatasetsac}, a large-scale collection of solved OPF instances derived from \texttt{PGLib-OPF} benchmark grids. Each instance is encoded as a heterogeneous graph in which the network’s physical backbone is represented by bus nodes connected through relations corresponding to AC lines and transformers.
The resulting representation supports learning tasks that map grid states and parameters to OPF-relevant quantities (e.g., operating points or feasibility/optimality surrogates), and \texttt{OPFData} additionally includes variants with topological perturbations to evaluate robustness under contingency-like structural changes, referred to as \texttt{N} (i.e., original in-distribution scenarios) and \texttt{N-1} scenarios (i.e., out of distribution scenarios where one line is disconnected).
Concretely, we evaluate on three benchmark topologies of increasing complexity: \texttt{case14} (14 buses), \texttt{case30} and \texttt{case118}.

\paragraph{Mean Squared Error.}
We measure the numerical accuracy of the predicted OPF solutions using the Mean Squared Error (MSE) between the model's predictions and the ground-truth solver outputs in the dataset. We report the sum of two MSE terms: the generator MSE, computed over the active and reactive power outputs for each generator, and the bus MSE, computed over the voltage magnitude and voltage angle at each bus. Beyond constraint satisfaction, measuring MSE ensures the solution remains numerically close from the optimal dispatch.

\paragraph{Evaluated Constraints.}
We assess each predicted OPF solution using three feasibility criteria. 
For each bus \(i\), we compute the active and the reactive power balance residuals
\(\Delta P_i = P_i^{\mathrm{spec}} - P_i^{\mathrm{imp}}\) and
\(\Delta Q_i = Q_i^{\mathrm{spec}} - Q_i^{\mathrm{imp}}\), where the specified injections are obtained from predicted generation minus demand, and the implied injections are those induced by the predicted voltage values under the instance's topology. A bus satisfies the power-flow constraints when:
$\sqrt{\Delta P_i^2 + \Delta Q_i^2} \le \varepsilon_{\mathrm{pf}}$. Voltage feasibility is defined by the bound: $v_i^{\min} \le \hat v_i \le v_i^{\max}$, and generator feasibility is defined by the active-power constraint: $P_g^{\min} \le \hat P_g \le P_g^{\max}$.
For each sample, we compute the fraction of buses and generators satisfying each of these conditions, and we report the dataset-level average of these per-sample fractions. In addition, we report an aggregate constraint satisfaction score obtained by micro-averaging all individual (bus and generator level) constraint checks across the full evaluation set, so that each checked constraint instance contributes equally to the final measure. Mathematically, let \(s\) indexes the evaluation samples, and denote by \(n_{\mathrm{pf}}^{(s)}\), \(n_{\mathrm{v}}^{(s)}\), and \(n_{\mathrm{g}}^{(s)}\) the number of evaluated buses and generators for the power-flow, voltage, and generator constraints, respectively. Let \(n_{\mathrm{pf,sat}}^{(s)}\), \(n_{\mathrm{v,sat}}^{(s)}\), and \(n_{\mathrm{g,sat}}^{(s)}\) denote the corresponding numbers of satisfied constraints. The final constraint score is defined as:
\begin{equation}
\label{eq:cs}
\mathrm{CS} =
100 \times
\frac{
\sum_{s}
\left(
n_{\mathrm{pf,sat}}^{(s)} +
n_{\mathrm{v,sat}}^{(s)} +
n_{\mathrm{g,sat}}^{(s)}
\right)
}{
\sum_{s}
\left(
n_{\mathrm{pf}}^{(s)} +
n_{\mathrm{v}}^{(s)} +
n_{\mathrm{g}}^{(s)}
\right)
}
\end{equation}

\section{Experimental Protocol}

\subsection{Models}

\paragraph{Model sizes.} We evaluate two model sizes to study the impact of scale and training regime on OPF reasoning: (i) small models with 8B parameters and (ii) medium models with approximately 32B parameters. The small models comprise \emph{Llama~3.1 8B} and \emph{Qwen3 8B} as baselines, alongside their \emph{DeepSeek-distilled} reasoning counterparts: \emph{Deepseek R1 Distill Llama 3.1 8B} and \emph{DeepSeek R1 Qwen3 8B}. The medium sizes comprise \emph{Qwen~2.5 32B} as a vanilla baseline and \emph{QwQ} (32B) as its reasoning-focused counterpart.

\subsection{Fine-tuning}

\paragraph{SFT.} To evaluate the impact of standard alignment in constraint satisfaction, we apply Supervised Fine-Tuning (SFT) to the vanilla instruct models. We use LoRA~\cite{hu2022lora} to train adapters on the training set of OPF problems. This process optimizes the model to predict the ground-truth solutions in zero-shot directly, ensuring the output format and basic physical constraints are learned through direct imitation of high-quality examples.

\paragraph{GRPO.} Reasoning variants are trained using Group Relative Policy Optimization (GRPO) \cite{liu2024deepseek}, a reinforcement learning objective that aligns model preferences with structured validity and feasibility rewards. Unlike standard RLHF, GRPO eliminates the need for a separate critic model, by sampling a group of multiple outputs for each prompt and using the relative reward of each output compared to the group average to compute the policy gradient. This allows the model to reason \textit{through} constraints iteratively. The reward $R$ decomposes into a structural component and a constraint-satisfaction component:
\[
R = r_{\mathrm{json}} + r_{\mathrm{CS}}
\]
The term $r_{\mathrm{json}} \in \{0, 0.5\}$ rewards adherence to the required JSON schema (0 for an unparsable output; 0.5 for a well-formed output). The term $r_{\mathrm{CS}} \in [0, 1]$ captures feasibility by measuring the fraction of evaluated constraints satisfied by the proposed solution as defined in equation \ref{eq:cs}.

\subsection{Evaluation} Vanilla models are assessed using in-context learning (ICL). We provide the same prompt templates and demonstration format across models, including fine-tuned ones.
Across all configurations, models are instructed to return outputs in a standardized JSON schema that encodes the candidate dispatch, key feasibility indicators, and any auxiliary reasoning traces permitted by the prompt.

At inference time, we request the same JSON schema used during training and parse model outputs with strict validation. For ICL conditions, we reuse the same samples across models and capacity tiers to ensure comparability. For zero-shot conditions, no pre-samples are provided and the prompt only specifies the schema, objectives, and any required units or conventions. For every experiment, the model is assessed on 1,000 different conversations, the last message of each being the query input. In all cases, performance is computed on the parsed outputs; malformed outputs can't be parsed and are not evaluated.

\section{Results}

\begin{figure*}[!ht]
    \centering
    \includegraphics[width=\linewidth]{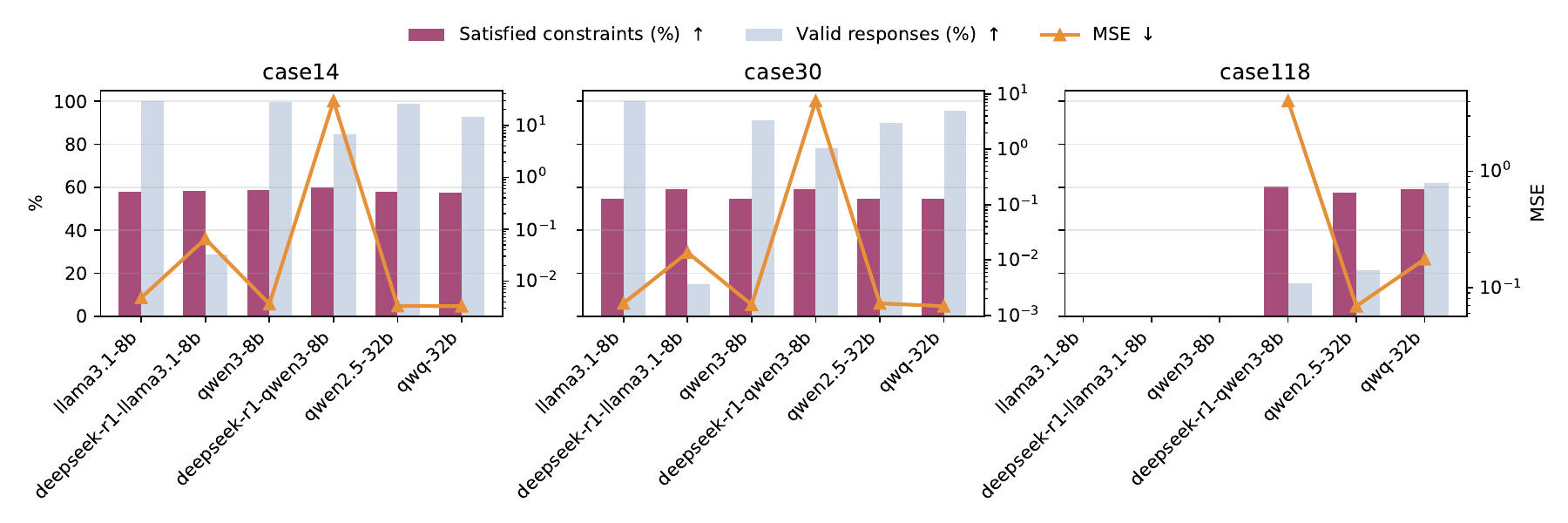}
    \caption{Performance of vanilla models (before fine-tuning) on cases 14, 30 and 118.}
    \label{fig:vanilla}
\end{figure*}

We report the performance of models on the three OPF benchmark cases (14, 30, 118), evaluated across three metrics: satisfied constraints, defined in equation \ref{eq:cs} (\%, higher is better), valid responses (\%, higher is better), and total MSE (lower is better). The table detailing the satisfied constraints decomposition is available in the data appendix.

\subsection{Vanilla models}

Figure~\ref{fig:vanilla} reports that across all three cases, the constraint satisfaction rate remains globally flat, hovering around 55--60\% for every model regardless of architecture or training regime.
This plateau is explained by the large non satisfaction of power flow constraints for all models, representing $\sim$40\% of all the evaluated constraints; suggesting that models are not actively reasoning through the power-flow equations but are instead producing outputs that satisfy a fixed subset of constraints by default,  likely through pattern matching rather than genuine optimization.

\paragraph{Non-reasoning vs.\ reasoning models.} 
Reasoning-augmented models do \textit{not} systematically improve constraint satisfaction over their vanilla counterparts. 
In several settings, reasoning models even degrade performance on the MSE axis: for instance, DeepSeek-R1-Qwen3-8B shows a sharp MSE spike on case14 and case30 (reaching values above 
$10^{1}$), indicating that extended chain-of-thought generation may lead to numerically unstable outputs. The only consistent advantage of reasoning models is a higher rate of 
\textit{valid} (parsable) responses in some configurations, yet this benefit is not universal.

\paragraph{Effect of model size.}
Scaling from 8B to 32B parameters yields limited gains on the constraint satisfaction score, which remains near 55--60\% for both Qwen2.5-32B and QwQ-32B, matching the
8B models. 
The most visible size-related effect is on response validity: larger models tend to produce well-formed JSON outputs more reliably, particularly on the larger case118 topology where smaller models completely fail in generating any parsable response. 
Regarding the MSE: although Qwq-32b achieves competitive MSE on cases 14 and 30, the performance doesn't improve monotically with scale.
These results suggest that simply scaling model size is insufficient to solve complex optimization like OPF.
\vspace{-5pt}

\subsection{Supervised Fine-Tuning}

\begin{figure}[!ht]
    \centering
    \includegraphics[width=\linewidth]{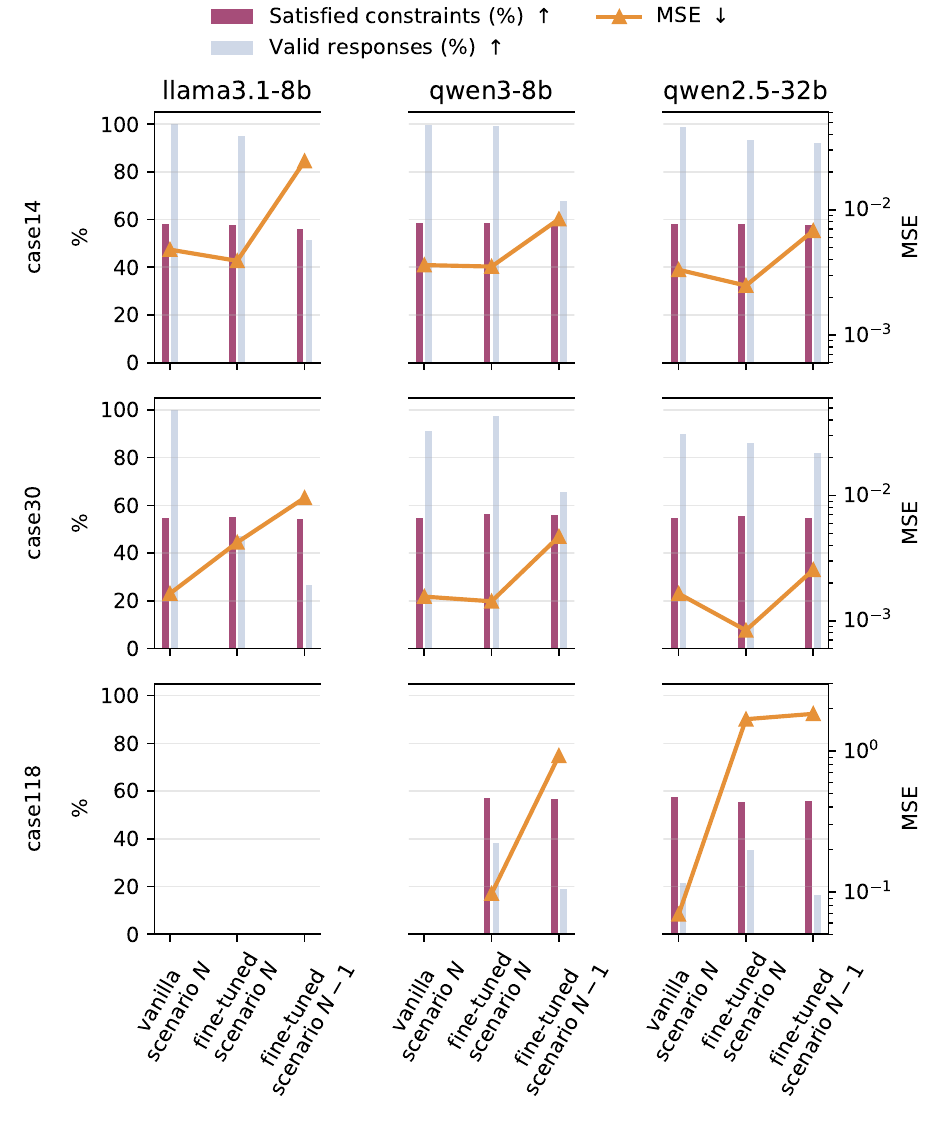}
    \caption{Performance of vanilla models (before fine-tuning) and of fine-tuned models on N and N-1 cases.}
    \label{fig:sft}
\end{figure}

\paragraph{Performance on scenario N.}
Figure \ref{fig:sft} shows that fine-tuning consistently improves response validity and MSE across all models and cases. Constraint satisfaction, however, remains largely unchanged after SFT at around 55--60\%, indicating that the models learn to produce \textit{well-formatted} but not necessarily \textit{physically feasible} solutions, confirming this \textit{reasoning shortcut} established in the state of the art.
\vspace{-3pt}

\paragraph{Scenario N vs. N-1.}
Under the $N-1$ contingency setting, valid response rates unsurprisingly drop noticeably for all grids, and MSE increases across all models, indicating that topological perturbations expose the limits of SFT-based generalization. Constraint satisfaction scores remain at the same plateau as in the vanilla and $N$ settings, confirming that simple fine-tuning on nominal cases does not confer robustness to structural grid changes.
\vspace{-3pt}

\subsection{Impact of GRPO}

\begin{table}
\centering
\footnotesize
\setlength{\tabcolsep}{5pt}
\def\arraystretch{1}
\begin{tabular}{p{7mm} p{5mm} l rrr}
\toprule
\textbf{Model} & \textbf{Case} & \textbf{Scenario} & \makecell{\textbf{Gen.} \\ \textbf{limits}}  & \makecell{\textbf{Voltage} \\ \textbf{limits}} & \textbf{PF} \\
\midrule
\midrule
  \multirow{6}{*}{\rotatebox{90}{\makecell{\textbf{Deepseek-R1} \\ \textbf{Qwen3-8B}}}} & \multirow{3}{*}{14} & Vanilla $N$ & 100\% & 98.7\% & 7.5\% \\
   &  & GRPO $N$ & 100\% & 99.6\% & 6.25\% \\
   &  & GRPO $N-1$ & 100\% & 98.6\% & 3.19\% \\
  \cmidrule(l){2-6}
   & \multirow{3}{*}{30} & Vanilla $N$ & 100\% & 98.9\% & 11.7\% \\
   &  & GRPO $N$ & 100\% & 99.5\% & 12.5\% \\
   &  & GRPO $N-1$ & 99.8\% & 97\% & 9.71\% \\
\midrule
  \multirow{6}{*}{\rotatebox{90}{\makecell{\textbf{Deepseek-R1} \\ \textbf{Llama3.1-8B}}}} & \multirow{3}{*}{14} & Vanilla $N$ & 99.9\% & 98.4\% & 3.94\% \\
   &  & GRPO $N$ & 100\% & 99.9\% & 1.37\% \\
   &  & GRPO $N-1$ & 99.9\% & 99.8\% & 0.83\% \\
  \cmidrule(l){2-6}
   & \multirow{3}{*}{30} & Vanilla $N$ & 100\% & 99.3\% & 10.6\% \\
   &  & GRPO $N$ & 100\% & 99.6\% & 24.6\% \\
   &  & GRPO $N-1$ & 100\% & 100\% & 17.8\% \\
\bottomrule
\end{tabular}
\caption{\% of satisfied constraints by model, case, and scenario.}
\label{tab:reasoning-constraints}
\end{table}

\begin{figure}[!ht]
    \centering
    \includegraphics[width=\linewidth]{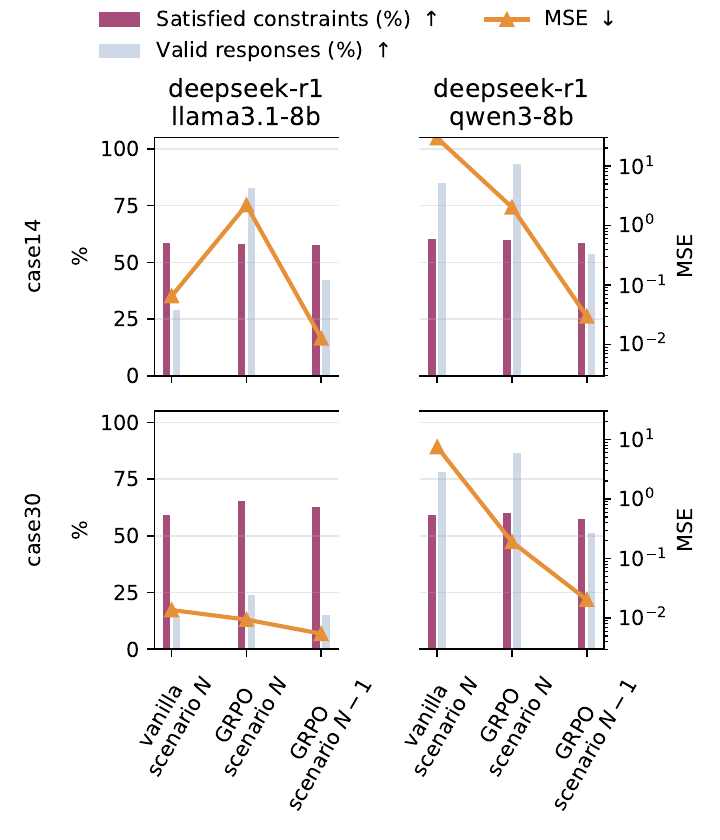}
    \caption{Performance of reasoning models, before GRPO (vanilla) and after GRPO, on N and N-1 scenarios.}
    \label{fig:grpo}
\end{figure}

\begin{table*}
  \centering
  \footnotesize
  \begin{tabular}{llrrrrrr}
    \toprule
    \textbf{Case} & \textbf{Model} & \multicolumn{3}{c}{in-context} & \multicolumn{3}{c}{zero-shot} \\
    \cmidrule(lr){3-5} \cmidrule(lr){6-8}
    & & \textbf{Valid} (\%) & \textbf{CS} (\%) $\uparrow$ & \textbf{MSE} $\downarrow$ & \textbf{Valid} (\%) & \textbf{CS} (\%) $\uparrow$ & \textbf{MSE} $\downarrow$ \\
    \midrule
    \texttt{case14} & llama3.1-8b & 95.0 & 57.8 & 3.91e-03 & 99.7 & 63.8 & 5.26e-03 \\
     & qwen3-8b & 99.3 & 58.4 & 3.52e-03 & 32.8 & 59.5 & 2.94e-02 \\
     & qwen2.5-32b & 93.5 & 58.2 & 2.48e-03 & 78.8 & 64.7 & 5.40e-03 \\
    \texttt{case30} & llama3.1-8b & 47.1 & 55.1 & 4.22e-03 & 82.5 & 61.9 & 2.60e-02 \\
     & qwen3-8b & 97.5 & 56.5 & 1.43e-03 & 35.6 & 55.1 & 2.46e-02 \\
     & qwen2.5-32b & 86.2 & 55.7 & 8.37e-04 & 96.4 & 59.8 & 1.01e-02 \\
    \texttt{case118} & llama3.1-8b & 0.0 & -- & -- & 0.0 & -- & -- \\
     & qwen3-8b & 38.3 & 57.2 & 9.71e-02 & 0.0 & -- & -- \\
     & qwen2.5-32b & 35.3 & 55.6 & 1.68e+00 & 0.0 & -- & -- \\
    \bottomrule
  \end{tabular}
  \caption{Performance of fine-tuned models, in-context learning vs. zero-shot inference.}
  \label{tab:zero-shot}
\end{table*}

\begin{table*}
  \centering
  \footnotesize
  \begin{tabular}{llrrrrrr}
    \toprule
    \textbf{Case} & \textbf{Model} & \multicolumn{3}{c}{basic} & \multicolumn{3}{c}{enhanced} \\
    \cmidrule(lr){3-5} \cmidrule(lr){6-8}
    & & \textbf{Valid} (\%) & \textbf{CS} (\%) $\uparrow$ & \textbf{MSE} $\downarrow$ & \textbf{Valid} (\%) & \textbf{CS} (\%) $\uparrow$ & \textbf{MSE} $\downarrow$ \\
    \midrule
    case14 & llama3.1-8b & 100.0 & 58.0 & 4.80e-03 & 100.0 & 58.0 & 4.92e-03 \\
     & qwen3-8b & 99.8 & 58.8 & 3.61e-03 & 99.6 & 59.0 & 3.61e-03 \\
    case30 & llama3.1-8b & 100.0 & 54.6 & 1.65e-03 & 99.9 & 55.4 & 1.57e-03 \\
     & qwen3-8b & 91.2 & 54.7 & 1.56e-03 & 98.5 & 60.4 & 1.66e-03 \\
    \bottomrule
  \end{tabular}
  \caption{Performance of inference with the \textit{basic} system prompt defined above, and inference with an LLM-enhanced system prompt.}
  \label{tab:enhanced-prompt}
\end{table*}

Figure~\ref{fig:grpo} reports the performance of reasoning models before and after GRPO fine-tuning on the $N$ and $N-1$ test scenarios.
GRPO training yields mixed results: while it improves response validity (can be attributed to the incorporation of the correct format reward), constraint satisfaction still mainly remains at near 55--60\%, except for the case 30, clearly getting over the 60\% bar for the first time, including for the $N-1$ contingency setting. Table \ref{tab:reasoning-constraints} especially shows that GRPO improves the power flow equations constraint on case 30 with a significant increase (+132\%) for Deepseek-R1-Llama3.1-8B, but degrades it on case 14.

\subsection{Ablations}
Table~\ref{tab:zero-shot} compares fine-tuned models evaluated in their training regime (zero-shot) against the same models prompted with in-context learning examples (ICL).
Across all models and grid sizes, ICL consistently yields lower MSE and higher valid response rates than zero-shot inference, suggesting that in-context examples provide structural guidance that compensates for the limitations of imitation-based fine-tuning, even when the model has already been exposed to the output format during training.

\paragraph{Impact of prompt}
To assess whether prompt quality influences model performance, we designed an enhanced version of the original system prompt using ChatGPT, enriching it with more explicit physical context, clearer formatting instructions, emphasizing on constraints and being domain-specific terminology related to power systems and OPF. As shown in Figure~\ref{tab:enhanced-prompt}, this enhancement has virtually no impact: all metrics remain essentially identical. The performance bottleneck might therefore not lie in the clarity or richness of the prompt formulation, but rather in the models' fundamental inability to reason over power-flow constraint.

\paragraph{Uncertainty and Calibration.}
To complement constraint satisfaction and MSE, we evaluate the calibration of constraint satisfaction predictions using Expected Calibration Error (ECE), Brier Score, and Negative Log-Likelihood (NLL) on Llama 3.1 8B for \texttt{case14}.
The vanilla model already exhibits poor calibration (ECE = 0.696, Brier = 0.485, NLL = 1.192), and SFT does not improve it;
scores slightly worsen on scenario $N$ (ECE = 0.720, Brier = 0.519, NLL = 1.275).
Interestingly, SFT on the $N-1$ out-of-distribution scenario yields marginally better calibration (ECE = 0.677, Brier = 0.458, NLL = 1.131) than the in-distribution counterpart, suggesting that the model's confidence is not meaningfully shaped by fine-tuning.

\section*{Conclusion}

We introduced a rigorous evaluation framework for assessing whether LLMs can reason and optimize under the physical and operational constraints of the Optimal Power Flow problem: a task demanding structured input interpretation, multi-step arithmetic, and simultaneous satisfaction of interacting physical constraints. Our empirical study yielded three consistent findings. All models, regardless of architecture or scale, show a plateau in constraint satisfaction, suggesting pattern matching rather than genuine optimization. Supervised fine-tuning reliably improves response formatting and marginally reduces MSE, but leaves constraint satisfaction essentially unchanged, confirming the shortcut-learning behavior reported in prior work. Reinforcement learning with constraint-satisfaction rewards yields modest but meaningful improvements on some topologies, leading some models exceed the constraints plateau, including in out-of-distribution examples.

Across all settings, no single model dominates on both metrics simultaneously. Qwen-family models tend to produce lower MSE and better-formed outputs, while Llama-based models show more stable constraint satisfaction under GRPO fine-tuning. Scaling to 32B parameters improves response validity on larger grids but does not resolve the constraint satisfaction plateau.

Future work should include more datasets from other domains, as well as richer reward signals that decompose individual constraint violations to guide more targeted reinforcement learning.

\clearpage

\section{Limitations}
\subsection{The "Memorization" vs. "Optimization" trap}

Since LLMs cannot actually perform 10 iterations of Newton-Raphson in their "latent space," they often resort to Result Guessing. If the model recognizes the input as "similar to a standard power grid, or if they recognize similar financial or cyber-security datasets, it will provide values that look like a valid solution. Our results using the N-1 out of distribution test set demonstrate an increased error compared to the N case (in-distribution) that suggest that even under fine-tuning and GRPO, LLMs mainly rely on memorization mechanisms.

\subsection{The lack of tools}

Using an LLM to evaluate OPF without tools is actually testing its calculating limitations, not only its logical reasoning. One could argue that a better paradigm is to restrict LLMs to abstraction tasks only: Reading the grid data, understanding the physics, deducing the correct constraints, and writing the mathematical formulation into solver code (like Pyomo or JuMP). Then leaving the solver (Gurobi, Ipopt) execute the heavy numerical matrix arithmetic and return the result. However, multiple research ave demonstrated the capabilities of deep learning models in reliable OPF predictions (our OPF papers), and a previous study suggested that LLM could also solve the OPF problem with small grids. In addition, our results suggest that not only the predicted solutions by LLMs exhibits high MSE, but even the understanding (valid outputs) and the basic constraint satisfaction are not supported.

\section*{Acknowledgements}
This work was partially funded by FNR CORE project LEAP (17042283) and by Creos Luxembourg S.A. \\
Dr. Ghamizi is supported by the Luxembourg National Research Fund (FNR) CORE C24/IS/18942843.

\section*{Ethical Considerations}
This work studies the ability of LLMs to reason under physical and operational constraints in optimal power flow settings, with the goal of better understanding their current limitations rather than advocating for their autonomous deployment in real-world power systems. Since power grid operation is extremely safety-critical, one important ethical implication of our findings is that LLM outputs should not be treated as reliable decision support mechanisms yet. However, at the same time our work may have positive societal value by providing a challenging benhcmark that helps identify failure modes and motivate safer hybrid systems combining LLMs with formal solvers and human oversight. More broadly, we hope that this work encourages future research on trustworthy reasoing an robustness for LLM-based systems deployed in critical application domains.

\clearpage



\begin{table*}
\centering
\label{tab:opf_results}
\footnotesize
\begin{tabular}{ll l rrr}
\toprule
\textbf{Model} & \textbf{Case} & \textbf{Scenario} & \makecell{\textbf{Gen.} \\ \textbf{limits}}  & \makecell{\textbf{Voltage} \\ \textbf{limits}} & \textbf{PF} \\
\midrule
  \multirow{2}{*}{\rotatebox{90}{\makecell{\textbf{QwQ} \\ \textbf{32B}}}} & \multirow{1}{*}{14} & Vanilla $N$ & 100\% & 100\% & 0.415\% \\
  \cmidrule(l){2-6}
   & \multirow{1}{*}{30} & Vanilla $N$ & 100\% & 100\% & 0.199\% \\
\midrule
  \multirow{6}{*}{\rotatebox{90}{\textbf{Qwen2.5-32B}}} & \multirow{3}{*}{14} & Vanilla $N$ & 100\% & 100\% & 1.19\% \\
   &  & SFT $N$ & 100\% & 100\% & 1.41\% \\
   &  & SFT $N-1$ & 100\% & 100\% & 0.426\% \\
  \cmidrule(l){2-6}
   & \multirow{3}{*}{30} & Vanilla $N$ & 100\% & 100\% & 0.134\% \\
   &  & SFT $N$ & 100\% & 100\% & 2.51\% \\
   &  & SFT $N-1$ & 100\% & 100\% & 0.747\% \\
\midrule
  \multirow{6}{*}{\rotatebox{90}{\textbf{LLaMA 3.1-8B}}} & \multirow{3}{*}{14} & Vanilla $N$ & 100\% & 100\% & 1\% \\
   &  & SFT $N$ & 100\% & 99.9\% & 0.609\% \\
   &  & SFT $N-1$ & 99.4\% & 96.4\% & 0.208\% \\
  \cmidrule(l){2-6}
   & \multirow{3}{*}{30} & Vanilla $N$ & 100\% & 100\% & 0.217\% \\
   &  & SFT $N$ & 100\% & 98.8\% & 2.35\% \\
   &  & SFT $N-1$ & 100\% & 98.8\% & 0.774\% \\
\midrule
  \multirow{6}{*}{\rotatebox{90}{\textbf{Qwen3-8B}}} & \multirow{3}{*}{14} & Vanilla $N$ & 100\% & 100\% & 2.77\% \\
   &  & SFT $N$ & 100\% & 100\% & 1.92\% \\
   &  & SFT $N-1$ & 100\% & 100\% & 0.609\% \\
  \cmidrule(l){2-6}
   & \multirow{3}{*}{30} & Vanilla $N$ & 100\% & 100\% & 0.241\% \\
   &  & SFT $N$ & 100\% & 100\% & 4.22\% \\
   &  & SFT $N-1$ & 99.8\% & 100\% & 2.98\% \\
\midrule
  \multirow{6}{*}{\rotatebox{90}{\makecell{\textbf{Deepseek-R1} \\ \textbf{Qwen3-8B}}}} & \multirow{3}{*}{14} & Vanilla $N$ & 100\% & 98.7\% & 7.5\% \\
   &  & GRPO $N$ & 100\% & 99.6\% & 6.25\% \\
   &  & GRPO $N-1$ & 100\% & 98.6\% & 3.19\% \\
  \cmidrule(l){2-6}
   & \multirow{3}{*}{30} & Vanilla $N$ & 100\% & 98.9\% & 11.7\% \\
   &  & GRPO $N$ & 100\% & 99.5\% & 12.5\% \\
   &  & GRPO $N-1$ & 99.8\% & 97\% & 9.71\% \\
\midrule
  \multirow{6}{*}{\rotatebox{90}{\makecell{\textbf{Deepseek-R1} \\ \textbf{Llama3.1-8B}}}} & \multirow{3}{*}{14} & Vanilla $N$ & 99.9\% & 98.4\% & 3.94\% \\
   &  & GRPO $N$ & 100\% & 99.9\% & 1.37\% \\
   &  & GRPO $N-1$ & 99.9\% & 99.8\% & 0.829\% \\
  \cmidrule(l){2-6}
   & \multirow{3}{*}{30} & Vanilla $N$ & 100\% & 99.3\% & 10.6\% \\
   &  & GRPO $N$ & 100\% & 99.6\% & 24.6\% \\
   &  & GRPO $N-1$ & 100\% & 100\% & 17.8\% \\
\bottomrule
\end{tabular}
\caption{\% of satisfied constraints by model, case, and scenario}
\end{table*}


\begin{figure*}
    \centering
    \includegraphics[width=0.6\linewidth]{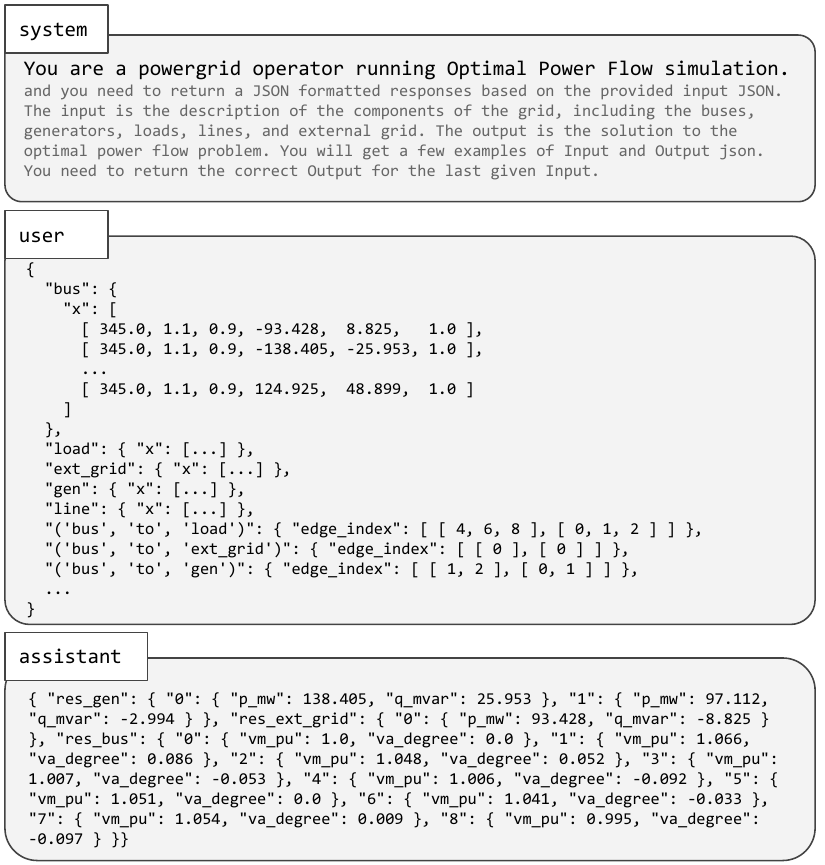}
    \caption{Prompt used for the OPF task assessment.}
    \label{fig:opf-example}
\end{figure*}


\begin{thebibliography}{30}
\providecommand{\natexlab}[1]{#1}

\bibitem[{Bernier et~al.(2025)Bernier, Cao, Cordy, and Ghamizi}]{bernier2025powergraph}
Fabien Bernier, Jun Cao, Maxime Cordy, and Salah Ghamizi. 2025.
\newblock Powergraph-llm: Novel power grid graph embedding and optimization with large language models.
\newblock \emph{IEEE Transactions on Power Systems}.

\bibitem[{Chen et~al.(2025)Chen, Wei, Ren, Li, and Zhang}]{chen2025lr2bench}
Jianghao Chen, Zhenlin Wei, Zhenjiang Ren, Ziyong Li, and Jiajun Zhang. 2025.
\newblock Lr$^2$bench: Evaluating long-chain reflective reasoning capabilities of large language models via constraint satisfaction problems.
\newblock In \emph{Findings of the Association for Computational Linguistics: ACL 2025}, pages 6006--6032.

\bibitem[{Chollet(2019)}]{chollet2019measureintelligence}
François Chollet. 2019.
\newblock \href {https://arxiv.org/abs/1911.01547} {On the measure of intelligence}.
\newblock \emph{Preprint}, arXiv:1911.01547.

\bibitem[{Chu et~al.(2025)Chu, Zhai, Yang, Tong, Xie, Schuurmans, Le, Levine, and Ma}]{chu2025sft}
Tianzhe Chu, Yuexiang Zhai, Jihan Yang, Shengbang Tong, Saining Xie, Dale Schuurmans, Quoc~V Le, Sergey Levine, and Yi~Ma. 2025.
\newblock Sft memorizes, rl generalizes: A comparative study of foundation model post-training.
\newblock \emph{arXiv preprint arXiv:2501.17161}.

\bibitem[{Duchnowski et~al.(2025)Duchnowski, Pavlick, and Koller}]{duchnowski-etal-2025-knapsack}
Alex Duchnowski, Ellie Pavlick, and Alexander Koller. 2025.
\newblock \href {https://doi.org/10.18653/v1/2025.findings-emnlp.352} {A knapsack by any other name: Presentation impacts {LLM} performance on {NP}-hard problems}.
\newblock In \emph{Findings of the Association for Computational Linguistics: EMNLP 2025}, pages 6628--6651, Suzhou, China. Association for Computational Linguistics.

\bibitem[{Guo et~al.(2025)Guo, Yang, Zhang, Song, Wang, Zhu, Xu, Zhang, Ma, Bi et~al.}]{guo2025deepseek}
Daya Guo, Dejian Yang, Haowei Zhang, Junxiao Song, Peiyi Wang, Qihao Zhu, Runxin Xu, Ruoyu Zhang, Shirong Ma, Xiao Bi, and 1 others. 2025.
\newblock Deepseek-r1 incentivizes reasoning in llms through reinforcement learning.
\newblock \emph{Nature}, 645(8081):633--638.

\bibitem[{Hu et~al.(2022)Hu, Shen, Wallis, Allen-Zhu, Li, Wang, Wang, Chen et~al.}]{hu2022lora}
Edward~J Hu, Yelong Shen, Phillip Wallis, Zeyuan Allen-Zhu, Yuanzhi Li, Shean Wang, Liang Wang, Weizhu Chen, and 1 others. 2022.
\newblock Lora: Low-rank adaptation of large language models.
\newblock \emph{Iclr}, 1(2):3.

\bibitem[{Huang et~al.(2024)Huang, Li, Liu, Wang, and Chen}]{huang2024large}
Chenghao Huang, Siyang Li, Ruohong Liu, Hao Wang, and Yize Chen. 2024.
\newblock Large foundation models for power systems.
\newblock In \emph{2024 IEEE Power \& Energy Society General Meeting (PESGM)}, pages 1--5. IEEE.

\bibitem[{Lin et~al.(2025)Lin, Bras, Richardson, Sabharwal, Poovendran, Clark, and Choi}]{lin2025zebralogic}
Bill~Yuchen Lin, Ronan~Le Bras, Kyle Richardson, Ashish Sabharwal, Radha Poovendran, Peter Clark, and Yejin Choi. 2025.
\newblock Zebralogic: On the scaling limits of llms for logical reasoning.
\newblock \emph{arXiv preprint arXiv:2502.01100}.

\bibitem[{Liu et~al.(2024{\natexlab{a}})Liu, Feng, Xue, Wang, Wu, Lu, Zhao, Deng, Zhang, Ruan et~al.}]{liu2024deepseek}
Aixin Liu, Bei Feng, Bing Xue, Bingxuan Wang, Bochao Wu, Chengda Lu, Chenggang Zhao, Chengqi Deng, Chenyu Zhang, Chong Ruan, and 1 others. 2024{\natexlab{a}}.
\newblock Deepseek-v3 technical report.
\newblock \emph{arXiv preprint arXiv:2412.19437}.

\bibitem[{Liu et~al.(2024{\natexlab{b}})Liu, Bai, Wen, Wang, Liu, Liang, Zhao, and Dong}]{liu2024lfllm}
Guolong Liu, Yan Bai, Keen Wen, Xinlei Wang, Yanli Liu, Gaoqi Liang, Junhua Zhao, and Zhao~Yang Dong. 2024{\natexlab{b}}.
\newblock Lfllm: A large language model for load forecasting.
\newblock \emph{Authorea Preprints}.

\bibitem[{Lovett et~al.(2024)Lovett, Zgubic, Liguori, Madjiheurem, Tomlinson, Elster, Apps, Witherspoon, and Piloto}]{lovett2024opfdatalargescaledatasetsac}
Sean Lovett, Miha Zgubic, Sofia Liguori, Sephora Madjiheurem, Hamish Tomlinson, Sophie Elster, Chris Apps, Sims Witherspoon, and Luis Piloto. 2024.
\newblock \href {https://arxiv.org/abs/2406.07234} {Opfdata: Large-scale datasets for ac optimal power flow with topological perturbations}.
\newblock \emph{Preprint}, arXiv:2406.07234.

\bibitem[{Madusanka et~al.(2024)Madusanka, Pratt-Hartmann, and Batista-Navarro}]{madusanka2024natural}
Tharindu Madusanka, Ian Pratt-Hartmann, and Riza~Theresa Batista-Navarro. 2024.
\newblock Natural language satisfiability: Exploring the problem distribution and evaluating transformer-based language models.
\newblock In \emph{Proceedings of the 62nd Annual Meeting of the Association for Computational Linguistics (Volume 1: Long Papers)}, pages 15278--15294.

\bibitem[{Michailidis et~al.(2025)Michailidis, Tsouros, and Guns}]{michailidis2025cp}
Kostis Michailidis, Dimos Tsouros, and Tias Guns. 2025.
\newblock Cp-bench: Evaluating large language models for constraint modelling.
\newblock \emph{arXiv preprint arXiv:2506.06052}.

\bibitem[{Michailidis et~al.(2024)Michailidis, Tsouros, Guns, and Shaw}]{michailidis2024constraint}
Kostis Michailidis, Dimos Tsouros, Tias Guns, and P.~Shaw. 2024.
\newblock Constraint modelling with llms using in-context learning.
\newblock In \emph{30th International Conference on Principles and Practice of Constraint Programming (CP 2024)}, volume 307, pages 20:1--20:15. Schloss Dagstuhl--Leibniz-Zentrum für Informatik.

\bibitem[{Mirshekali et~al.(2025)Mirshekali, Shadi, Ladani, and Shaker}]{mirshekali2025review}
Hamid Mirshekali, Mohammad~Reza Shadi, Fatemehsadat~Ghanadi Ladani, and Hamid~Reza Shaker. 2025.
\newblock A review of large language models for energy systems: Applications, challenges, and future prospects.
\newblock \emph{IEEE Access}.

\bibitem[{Ni et~al.(2025)Ni, Chen, Li, Chen, Li, Wang, Wang, Wang, Zhang, Fan et~al.}]{ni2025survey}
Shiwen Ni, Guhong Chen, Shuaimin Li, Xuanang Chen, Siyi Li, Bingli Wang, Qiyao Wang, Xingjian Wang, Yifan Zhang, Liyang Fan, and 1 others. 2025.
\newblock A survey on large language model benchmarks.
\newblock \emph{arXiv preprint arXiv:2508.15361}.

\bibitem[{Pan et~al.(2023)Pan, Albalak, Wang, and Wang}]{pan2023logic}
Liangming Pan, Alon Albalak, Xinyi Wang, and William Wang. 2023.
\newblock Logic-lm: Empowering large language models with symbolic solvers for faithful logical reasoning.
\newblock In \emph{Findings of the Association for Computational Linguistics: EMNLP 2023}, pages 3806--3824.

\bibitem[{Qiu et~al.(2024)Qiu, Li, Wang, Xie, Zhang, Mo, Chen, and Dong}]{qiu2024ef}
Zihang Qiu, Chaojie Li, Zhongyang Wang, Renyou Xie, Borui Zhang, Huadong Mo, Guo Chen, and Zhaoyang Dong. 2024.
\newblock Ef-llm: Energy forecasting llm with ai-assisted automation, enhanced sparse prediction, hallucination detection.
\newblock \emph{arXiv preprint arXiv:2411.00852}.

\bibitem[{Ramamonjison et~al.(2023)Ramamonjison, Yu, Li, Li, Carenini, Ghaddar, He, Mostajabdaveh, Banitalebi-Dehkordi, Zhou et~al.}]{ramamonjison2023nl4opt}
Rindranirina Ramamonjison, Timothy Yu, Raymond Li, Haley Li, Giuseppe Carenini, Bissan Ghaddar, Shiqi He, Mahdi Mostajabdaveh, Amin Banitalebi-Dehkordi, Zirui Zhou, and 1 others. 2023.
\newblock Nl4opt competition: Formulating optimization problems based on their natural language descriptions.
\newblock In \emph{NeurIPS 2022 competition track}, pages 189--203. PMLR.

\bibitem[{Rein et~al.(2024)Rein, Hou, Stickland, Petty, Pang, Dirani, Michael, and Bowman}]{rein2024gpqa}
David Rein, Betty~Li Hou, Asa~Cooper Stickland, Jackson Petty, Richard~Yuanzhe Pang, Julien Dirani, Julian Michael, and Samuel~R Bowman. 2024.
\newblock Gpqa: A graduate-level google-proof q\&a benchmark.
\newblock In \emph{First conference on language modeling}.

\bibitem[{Ren et~al.(2025)Ren, Lai, Taylor, and Guo}]{ren2025can}
Xinxing Ren, Chun~Sing Lai, Gareth Taylor, and Zekun Guo. 2025.
\newblock Can large language model agents balance energy systems?
\newblock \emph{arXiv preprint arXiv:2502.10557}.

\bibitem[{Srivastava et~al.(2023)Srivastava, Rastogi, Rao, Shoeb, Abid, Fisch, Brown, Santoro, Gupta, Garriga-Alonso et~al.}]{srivastava2023beyond}
Aarohi Srivastava, Abhinav Rastogi, Abhishek Rao, Abu Awal~Md Shoeb, Abubakar Abid, Adam Fisch, Adam~R Brown, Adam Santoro, Aditya Gupta, Adri{\`a} Garriga-Alonso, and 1 others. 2023.
\newblock Beyond the imitation game: Quantifying and extrapolating the capabilities of language models.
\newblock \emph{Transactions on machine learning research}.

\bibitem[{Valmeekam et~al.(2023)Valmeekam, Marquez, Olmo, Sreedharan, and Kambhampati}]{valmeekam2023planbench}
Karthik Valmeekam, Matthew Marquez, Alberto Olmo, Sarath Sreedharan, and Subbarao Kambhampati. 2023.
\newblock Planbench: An extensible benchmark for evaluating large language models on planning and reasoning about change.
\newblock \emph{Advances in Neural Information Processing Systems}, 36:38975--38987.

\bibitem[{Voboril et~al.(2025)Voboril, Ramaswamy, and Szeider}]{voboril2025generating}
Florentina Voboril, Vaidyanathan~Peruvemba Ramaswamy, and Stefan Szeider. 2025.
\newblock Generating streamlining constraints with large language models.
\newblock \emph{Journal of Artificial Intelligence Research}, 84.

\bibitem[{Wang et~al.(2024)Wang, Ma, Zhang, Ni, Chandra, Guo, Ren, Arulraj, He, Jiang et~al.}]{wang2024mmlu}
Yubo Wang, Xueguang Ma, Ge~Zhang, Yuansheng Ni, Abhranil Chandra, Shiguang Guo, Weiming Ren, Aaran Arulraj, Xuan He, Ziyan Jiang, and 1 others. 2024.
\newblock Mmlu-pro: A more robust and challenging multi-task language understanding benchmark.
\newblock \emph{Advances in Neural Information Processing Systems}, 37:95266--95290.

\bibitem[{Wang(2024)}]{wang2024causalbench}
Zeyu Wang. 2024.
\newblock Causalbench: A comprehensive benchmark for evaluating causal reasoning capabilities of large language models.
\newblock In \emph{Proceedings of the 10th SIGHAN Workshop on Chinese Language Processing (SIGHAN-10)}, pages 143--151.

\bibitem[{Wei et~al.(2025)Wei, Wu, Wan, Suresh, Tan, Zhou, Koyejo, Wang, and Aiken}]{wei2025satbench}
Anjiang Wei, Yuheng Wu, Yingjia Wan, Tarun Suresh, Huanmi Tan, Zhanke Zhou, Sanmi Koyejo, Ke~Wang, and Alex Aiken. 2025.
\newblock Satbench: Benchmarking llms’ logical reasoning via automated puzzle generation from sat formulas.
\newblock In \emph{Proceedings of the 2025 Conference on Empirical Methods in Natural Language Processing}, pages 33820--33837.

\bibitem[{Yan et~al.(2024)Yan, Zhenyuan, Xu, and Zhou}]{yan2024probabilistic}
Ziming Yan, Du~Zhenyuan, Yan Xu, and Ziyan Zhou. 2024.
\newblock Probabilistic pv power forecasting by a multi-modal method using gpt-agent to interpret weather conditions.
\newblock In \emph{2024 IEEE 19th Conference on Industrial Electronics and Applications (ICIEA)}, pages 1--6. IEEE.

\bibitem[{Ye et~al.(2023)Ye, Chen, Dillig, and Durrett}]{ye2023satlm}
Xi~Ye, Qiaochu Chen, Isil Dillig, and Greg Durrett. 2023.
\newblock Satlm: Satisfiability-aided language models using declarative prompting.
\newblock \emph{Advances in Neural Information Processing Systems}, 36:45548--45580.

\end{thebibliography}
\end{document}